\documentclass[10pt,twocolumn,letterpaper]{article}

\usepackage[pagenumbers]{cvpr} % To force page numbers, e.g. for an arXiv version

\usepackage{graphicx}
\usepackage{amsmath}
\usepackage{amssymb}
\usepackage{booktabs}
\usepackage{makecell}
\usepackage{float}
\usepackage{physics}
\usepackage{nicefrac}

\usepackage[hyphens]{url}
\usepackage[pagebackref,breaklinks,colorlinks]{hyperref}
\usepackage[hyphenbreaks]{breakurl}
\usepackage[capitalize]{cleveref}
\crefname{section}{Sec.}{Secs.}
\Crefname{section}{Section}{Sections}
\Crefname{table}{Table}{Tables}
\crefname{table}{Tab.}{Tabs.}

\def\cvprPaperID{*****} % *** Enter the CVPR Paper ID here
\def\confName{CVPR}
\def\confYear{2022}

\begin{document}

%%%%%%%%% TITLE - PLEASE UPDATE
\title{Action Spotting using Dense Detection Anchors Revisited: Submission to the SoccerNet Challenge 2022}

\author{João V. B. Soares \qquad Avijit Shah\\
Yahoo Research\\
%Institution1 address\\
{\tt\small \{jvbsoares, avijit.shah\}@yahooinc.com}
% For a paper whose authors are all at the same institution,
% omit the following lines up until the closing ``}''.
% Additional authors and addresses can be added with ``\and'',
% just like the second author.
% To save space, use either the email address or home page, not both
%\and
%Second Author\\
%Institution2\\
%First line of institution2 address\\
%{\tt\small secondauthor@i2.org}
}
\maketitle

%%%%%%%%% ABSTRACT
\begin{abstract}
   This brief technical report describes our submission to the Action Spotting SoccerNet Challenge 2022. The challenge was part of the CVPR 2022 ActivityNet Workshop. Our submission was based on a recently proposed method which focuses on increasing temporal precision via a densely sampled set of detection anchors.
   %An anchor is defined as a pair formed by a timestamp and an action class.
   %Anchors are densely sampled, at a high temporal frequency. 
   %Each anchor has its own associated detection confidence and fine-grained temporal displacement.
   Due to its emphasis on temporal precision, this approach had shown significant improvements in the tight average-mAP metric. Tight average-mAP was used as the evaluation criterion for the challenge, and is defined using small temporal evaluation tolerances, thus being more sensitive to small temporal errors.
   %The approach using dense detection anchors is thus well poised to achieve a high tight average-mAP.
   In order to further improve results, here we introduce small changes in the pre- and post-processing steps, and also combine different input feature types via late fusion. These changes brought improvements that helped us achieve the first place in the challenge and also led to a new state-of-the-art on SoccerNet's test set when using the dataset's standard experimental protocol. This report briefly reviews the action spotting method based on dense detection anchors, then focuses on the modifications introduced for the challenge. We also describe the experimental protocols and training procedures we used, and finally present our results.
\end{abstract}

%%%%%%%%% BODY TEXT
\section{Introduction}
\label{sec:intro}

This brief report describes our submission to the action spotting track of the SoccerNet Challenge 2022, which came in first place in the track~\cite{mmsports, spottingchallenge}. In addition, when using SoccerNet's standard (non-challenge) experimental protocol, the same approach that was used for the submission sets a new state-of-the-art on the dataset.

Our submission was largely based on our recently proposed action spotting method~\cite{soares2022temporally}. The method relies on a dense set of detection anchors, with the goal of increasing the temporal precision of the detections. For each anchor, a detection confidence is inferred, along with a fine-grained temporal displacement, which indicates the time instant at which an action was predicted to happen. That method is briefly reviewed here in Section~\ref{sec:dense}. It consists of a two-phase approach, in which first a set of features is computed from the video frames, and then a model is applied on those features to infer which actions occur in the video and when. Section~\ref{sec:features} describes the features we experimented with, as well as the late fusion strategy that we used in order to combine different features together. Our features were based on feature sets that had been made available by previous authors~\cite{zhou2021feature,deliege2021soccernet}. Section~\ref{sec:training} describes our experimental protocols and training procedures. Section~\ref{sec:nms} describes a new post-processing method for action spotting, which is a simplified version of the Soft-NMS~\cite{bodla2017soft} method that was originally developed for post-processing object detection results. Finally, Section~\ref{sec:results} presents our experimental results.

\section{Action spotting via dense detection anchors}
\label{sec:dense}

For our submissions, we used the recent action spotting approach proposed by Soares et al.~\cite{soares2022temporally}. Inspired by dense single-stage object detectors~\cite{lin2017focal}, the approach uses a dense set of anchors to produce temporally precise detections in a single stage. Each anchor is defined as a pair formed by a time instant and an action class. Anchors are taken at regularly spaced time instants, at the same frequency as the input feature vectors, usually 1 or 2 per second, allowing for precise temporal localization. For each anchor, both a detection confidence and a fine-grained temporal displacement are inferred, with the temporal displacement indicating exactly when an action was predicted to happen, thus further increasing the precision of the results. Both types of outputs are then combined by displacing each confidence by its respective predicted temporal displacement. For the trunk of the model, here we experimented with a 1D version of the u-net, which was shown to be significantly faster than the Transformer Encoder alternative, while producing similar results~\cite{soares2022temporally}. The u-net is able to incorporate large temporal contexts that are important for action spotting, while at the same time preserving the smaller-scale features required for precise localization. Finally, a non-maximum suppression (NMS) step is applied to obtain the final detections. 

\section{Features}
\label{sec:features}

\subsection{Temporal resampling of features}
\label{sec:resample}

Our method uses a standard two-phase approach. The first phase decodes a video chunk and computes feature vectors representing each frame or block of frames. The feature vectors are then concatenated into a matrix, which represents the whole video chunk. The second phase of the approach then takes this feature matrix as input and produces the set of detected actions.

In order to produce good results, our experiments use the features provided by Zhou et al.~\cite{zhou2021feature}, which we refer to here as {\it Combination} features, given that they were produced by concatenating the features from a series of different fine-tuned models. These features were originally computed at 1 feature vector per second. Our method was implemented in such a way that the set of possible time instants for the final detections is the same as that of the extracted feature vectors. For the case of the Combination features, this results in possible output detections at every 1 second, following the same frequency as the input feature vectors. We noticed that this frequency was too low for appropriately computing the tight average-mAP metric, whose tolerance radius (equal to half the tolerance window size $\delta$) increases in half-second increments. 
%We thus needed a way to increase the output frequency to at least 2 per second. We achieved this by 
We thus resampled the Combination features using linear interpolation, to the desired frequency of 2 feature vectors per second. Using these resampled features also makes it easy to fuse them with other features that are already at the desired frequency of 2 feature vectors per second, as described below.

\subsection{Late fusion of different feature types}

%We initially set out to train a model on feature vectors formed by concatenating the Combination features, resampled to 2 per second as described above, with the ResNet features, which were already originally computed at a frequency of 2 per second. However, training on these features turned out to be fairly expensive in terms of both memory and compute. 

We also experimented with combining the resampled Combination features with the ResNet features from Deliege et al.~\cite{deliege2021soccernet}, which were already available at the desired frequency of 2 per second. As an easy way of combining these features, we adopted a late fusion approach, as follows. To obtain the fused detection confidences, we calculated a weighted average of the logits of the confidences from two previously trained models: one trained on the resampled Combination features, and the other trained on the ResNet features. The optimal weights were found through exhaustive search on the validation set. In a similar manner, we experimented with fusing the temporal displacements from models trained on the different feature types. However, this brought no improvement, so we opted to instead use just the single temporal displacement model trained on the resampled Combination features.

A simple alternative to late fusion would be early fusion implemented by concatenating the resampled Combination features with the ResNet features. This is an interesting direction to explore, though we avoided this approach since training on the concatenated features turned out to be fairly expensive in terms of memory and compute. Vanderplaetse and Dupont~\cite{vanderplaetse2020improved} experimented with combining ResNet and audio-based features using pooling-based architectures on the original SoccerNet dataset~\cite{giancola2018soccernet}. They showed that early fusion was not effective, while late fusion brought a significant improvement, and mid-level fusion provided the best overall results. At the same time, when Nerg{\aa}rd Rongved et al.~\cite{rongved2020real} experimented with combining visual and audio features on SoccerNet, they saw mixed results on whether early or late fusion should be preferred. Additionally, they report that combining the two modalities only provided clear improvements for the {\it Goal} class. Further investigation would be needed in order to understand how to best combine different types of features across different model architectures.

\begin{table*}
%\scriptsize
\footnotesize
%\small
  \centering
  \begin{tabular}{lllcccccccc}
    \toprule
     \makecell[l]{Experimental\\protocol} & 
    \makecell[l]{Post-\\processing} & Features & 
    \makecell{Validation\\tight a-mAP} & \makecell{Validation\\a-mAP} & 
    \makecell{Test tight\\a-mAP} & \makecell{Test\\a-mAP}& 
    \makecell{Challenge\\tight a-mAP} & \makecell{Challenge\\a-mAP}\\
    %\Xhline{2\arrayrulewidth}
    \midrule[2\arrayrulewidth]
    Test & NMS@20 & Combination & - & - & 60.7$^\dag$ & 77.3$^\dag$ & - & -\\
    \hline
    Test & NMS@20 & Combination$\times 2$ & 64.1 & 77.3 & 62.6 & 77.6 & - & -\\
    \hline
    Test & NMS$^*$ & Combination$\times 2$ & 65.1 & 76.6 & 64.0 & 76.8 & - & -\\
    \hline
    Test & Soft-NMS$^*$ & Combination$\times 2$ & 65.4 & 76.9 & 64.3 & 77.2 & - & -\\
    \hline
    Test & Soft-NMS$^*$ & Combination$\times 2$ ensemble & 66.2 & 77.5 & 64.6 & 77.5 & - & -\\
    \hline
    Test & Soft-NMS$^*$ & Combination$\times 2$ + ResNet & 66.3 & 78.2 & 65.1 & 78.5 & - & -\\
    \hline
    Challenge Validated & Soft-NMS$^*$ & Combination$\times 2$ & 65.8 & 76.5 & - & - & 65.1$^\ddagger$ & 75.9$^\ddagger$\\
    Challenge Validated & Soft-NMS$^*$ & Combination$\times 2$ & 65.6 & 76.3 & - & - & 64.6$^\ddagger$ & 74.9$^\ddagger$\\
    \hline
    Challenge & Soft-NMS$^*$ & Combination$\times 2$ & - & - & - & - & 67.0$^\ddagger$ & 77.3$^\ddagger$\\
    Challenge & Soft-NMS$^*$ & Combination$\times 2$ & - & - & - & - & 66.8$^\ddagger$ & 76.8$^\ddagger$\\
    \hline
    Challenge & Soft-NMS$^*$ & Combination$\times 2$ + ResNet & - & - & - & - & 67.6$^\ddagger$ & 77.9$^\ddagger$\\
    Challenge & Soft-NMS$^*$ & Combination$\times 2$ + ResNet & - & - & - & - & 67.8$^\ddagger$ & 78.0$^\ddagger$\\
    \bottomrule
  \end{tabular}
  \caption{Results of our method when varying: the protocols for training, validation, and testing; post-processing procedures; and input features. The results over the {\it Test} protocol are averages over five different training runs. On the other hand, for the protocols that present results on the challenge set ({\it Challenge Validated} and {\it Challenge}), we present results from individual runs, shown grouped into pairs, so that each setup takes up a pair of rows in the table. The difference between each of the paired runs is only in the random initialization of weights during training, leading to some variation in the results. ``Combination$\times 2$'' refers to the Combination features from Zhou et al.~\cite{zhou2021feature} resampled to 2 feature vectors per second, while ``Combination$\times 2$ + ResNet'' refers to our late fusion approach, and ``Combination$\times 2$ ensemble'' refers to applying late fusion to two models that are both trained on the resampled Combination features. *~Unless stated otherwise, NMS and Soft-NMS use the window size that maximizes tight average-mAP on the validation set. $\dag$~Numbers taken from our prior work~\cite{soares2022temporally}. $\ddagger$~Results provided by the evaluation server.}
  \label{tab:results}
\end{table*}

\section{Experimental protocol and model training}
\label{sec:training}

We use three different protocols in our experiments, each combining the original SoccerNet data splits in different ways. The dataset provides the following pre-determined set of splits: training, validation, test, and challenge. The challenge split labels are not provided with the dataset, so metrics on this split can only be obtained by submitting results to the evaluation server. The protocols that we experimented with are defined as follows. \begin{enumerate}
    \item The first protocol, named {\it Test}, trains on the training split, runs validation on the validation split, and tests on the test split. Our results on this protocol are presented as averages over five different training runs. This protocol does not involve the challenge split.
    \item The second protocol, named {\it Challenge Validated}, trains on {\it both} the training and test splits, runs validation on the validation split, and tests on the challenge split.
    \item The third protocol, named {\it Challenge}, trains on all the available labeled data (training, validation, and test splits), does not include any validation, and tests on the challenge split. The hyper-parameters for this last protocol are taken directly from those found using the {\it Challenge Validated} protocol above.
\end{enumerate}
For the {\it Challenge Validated} and {\it Challenge} protocols, we present results from individual runs instead of averages over multiple runs, since there is a limit on the number of challenge submissions that the evaluation server accepts, making it impractical to compute averages over multiple runs.

We follow the original training procedure for the models~\cite{soares2022temporally}. Of note, two different models are trained: one to predict detection confidences, and another to predict temporal displacements. 
The confidence and temporal displacement models each have 22.9M parameters when built on the resampled Combination features, and 18.5M when built on the ResNet features. For each model, the validation set is used in order to tune each of the following hyper-parameters in turn: the learning rate, $\rho$ for Sharpness-Aware Minimization (SAM)~\cite{foret2021sharpnessaware}, weight decay, and $\alpha$ for mixup data augmentation~\cite{zhang2018mixup}, with mixup only being applied to the model that predicts detection confidences.

% Baidu+ResNet: 22,905,297
% Baidu only: 21,856,721
% ResNet only: 18,514,385

\section{Soft non-maximum suppression}
\label{sec:nms}

To post-process the detection results, we applied a simplified one-dimensional version of Soft Non-Maximum Suppression (Soft-NMS)~\cite{bodla2017soft}. Soft-NMS was originally proposed for post-processing object detection results, as a generalization of the well-known NMS algorithm. Whereas standard NMS will remove any detections that have a high overlap with an accepted high-confidence detection, Soft-NMS opts to instead only {\it decay} the confidences associated with those overlapping detections, with the decay being proportional to the overlap. To adapt this to the one-dimensional action spotting setting, we define the decay as being proportional to the temporal distance between detections. More specifically, given an accepted high-confidence detection at time $t$, for any detection of the same class at time $s$, we define the decay function as $f(s) = \min\{\abs{s - t} / (\nicefrac{w}{2}), 1\}$, where $w$ is a window size that we choose on the validation set. The confidences are then updated by multiplication with the decay function $f$.

Whereas previous works focused on the average-mAP metric and found the optimal NMS window size to be around 20 seconds~\cite{giancola2021temporally, zhou2021feature, soares2022temporally}, when optimizing for tight average-mAP, we generally find that much smaller window sizes perform better. %Due to the tight average-mAP metric being used, we found it beneficial to tweak the NMS window size that is applied as a post-processing step.
More specifically, when optimizing tight average-mAP over the validation set, for regular NMS we found that a window size of 3 seconds usually produced the best results, while for Soft-NMS we found a window size of 8 seconds to be best.

\section{Results}
\label{sec:results}

%All the results presented here were obtained using Soft-NMS. Generally, applying Soft-NMS gives a very small but consistent improvement relative to standard NMS. For example, when using the {\it Challenge Validated} protocol, on the validation set, one of our models produced 65.78 tight average-mAP using Soft-NMS, versus 65.46 when using standard NMS, where each result corresponds to its respective optimal window size.

Table~\ref{tab:results} presents our results under different setups. Results that use the {\it Test} protocol are shown in the top rows of the table. For comparison purposes, we include the best results from our previous work~\cite{soares2022temporally}, which used the {\it Test} protocol with the original Combination features~\cite{zhou2021feature} and an NMS window size of 20 seconds. As shown in the table, by using the same approach, but resampling the Combination features to 2 per second (which also results in a u-net with more parameters), the tight average-mAP goes from 60.7 to 62.6. This was mainly due to increasing the frequency of the model outputs so that they match the successive increments of the evaluation tolerance radius, as explained in Section~\ref{sec:resample}.

The table also shows that by choosing the NMS window size as to optimize the validation tight average-mAP, we further improve tight average-mAP, but at the cost of decreasing the standard average-mAP. When applying Soft-NMS instead of NMS (with both approaches using their respective window sizes optimized for validation tight average-mAP), we obtain small improvements in both tight average-mAP and standard average-mAP. While the improvements from Soft-NMS are very small, we found them to be consistent across several other experiments (not reported here).

We then experiment with adding the ResNet features to the resampled Combination features using our simple late fusion approach, resulting in our best performance under the {\it Test} protocol: 65.1 tight average-mAP and 78.5 standard average-mAP. As a control experiment, we also apply late fusion to pairs of models that were both trained on the same resampled Combination features, which we denote as ``Combination$\times 2$ ensemble.'' As shown in the table, this ensemble approach provides a very small benefit, whereas adding the ResNet features to the resampled Combination features provides a larger improvement. This indicates that the two different features types (Combination and ResNet) are indeed complementary.

Table~\ref{tab:results} also presents our results on the challenge split, which were obtained from the challenge evaluation server. We see significant improvements when switching experimental protocols from {\it Challenge Validated} to {\it Challenge}, due to the increase in available training data. When we use late fusion to add the ResNet features to the resampled Combination features, we again see an improvement, leading to the our best submitted challenge results: 67.8 tight average-mAP and 78.0 standard average-mAP.

Table~\ref{tab:testresults} compares our results against those of published methods using the {\it Test} protocol, showing that our approach sets a new state-of-the-art.

The final results of the challenge, along with a brief description of many of the methods used in the submissions, will be made available by the challenge organizers~\cite{mmsports, spottingchallenge}.

\begin{table}
%\scriptsize
\footnotesize
%\small
  \centering
  \begin{tabular}{cccc}
    \toprule
    Method & Features & 
    \makecell{Test tight\\a-mAP} & \makecell{Test\\a-mAP}\\
    %\Xhline{2\arrayrulewidth}
    \midrule[2\arrayrulewidth]
    %\hline
    CALF~\cite{deliege2021soccernet,cioppa2020context} & ResNet+PCA & 12.2$^\dag$ & 40.7 \\
    \hline
    CALF-calibration~\cite{cioppa2021camera} & \makecell{ResNet+PCA\\+ top view} & N/A & 46.8\\
    \hline
    % CALF tight should be around 14.10%, from website.
    %NetVLAD++~\cite{giancola2021temporally} & ResNet+PCA & 11.3$^\dag$ &  50.7 \\
    %DU+SAM+mixup~\cite{soares2022temporally} & ResNet+PCA & 50.7 & 72.2 \\
    NetVLAD++~\cite{giancola2021temporally} & ResNet & 11.5$^*$ & 53.4 \\ 
    \hline
    AImageLab RMSNet~\cite{tomei2021rms} & ResNet tuned & 28.8$^*$ & 63.5$^*$ \\
    \hline
    Zhou et al.~\cite{zhou2021feature} & Combination &  47.1$^*$ & 73.8 \\
    \hline
    Faster-TAD~\cite{chen2022faster} & Swin tuned & 54.1 & N/A \\
    \hline
    E2E-Spot~\cite{hong2022spotting} & \makecell{RegNetY tuned} & 61.8 & 74.1\\
    \hline
    DU+SAM+mixup~\cite{soares2022temporally} & Combination & 60.7 & 77.3 \\
    \hline
    \makecell{DU+SAM+mixup\\+Soft-NMS (ours)} & \makecell{Combination$\times 2$\\+ ResNet} & 65.1 & 78.5 \\
    %\hline
    %\hline
    %\Xhline{2\arrayrulewidth}
    \bottomrule
  \end{tabular}
  \caption{Comparison with results from prior works using the {\it Test} protocol.  ``Combination$\times 2$ + ResNet'' refers to our late fusion approach, using the Combination features from Zhou et al.~\cite{zhou2021feature} resampled to 2 feature vectors per second, along with ResNet features. *~Results reported on challenge website~\cite{spottingchallenge}. $\dag$~Results computed using the implementation provided by the authors. }
  \label{tab:testresults}
\end{table}

\vspace{4mm}

\noindent {\bf Acknowledgements} We are grateful to Topojoy Biswas and Gaurav Srivastava for insightful discussions and feedback.

%\clearpage

%%%%%%%%% REFERENCES
{\small
\bibliographystyle{ieee_fullname}
\bibliography{references}
}

\end{document}